\documentclass{article}

\usepackage{PRIMEarxiv}

\usepackage[utf8]{inputenc} 
\usepackage[T1]{fontenc}    
\usepackage{hyperref}       
\usepackage{url}            
\usepackage{booktabs}       
\usepackage{amsfonts}       
\usepackage{nicefrac}       
\usepackage{microtype}      
\usepackage{lipsum}
\usepackage{graphicx}
\graphicspath{ {./images/} }

\makeatletter
\let\cas@beginabstract\abstract      
\let\cas@endabstract  \endabstract   
\makeatother

\usepackage{arabtex}
\usepackage{utf8}
\setcode{utf8}

\makeatletter
\let\abstract   \cas@beginabstract   
\let\endabstract\cas@endabstract     
\makeatother

\title{
  \begin{minipage}{0.20\textwidth}
    \includegraphics[height=2.2cm]{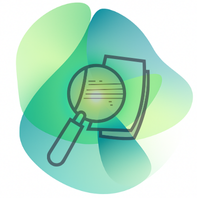}
  \end{minipage}
  \begin{minipage}{0.79\textwidth}
  \centering
    UI-Level Evaluation of ALLaM 34B:\\
    Measuring an Arabic-Centric LLM via HUMAIN Chat
  \end{minipage}%
}

\author{
 Omer Nacar \\
 NAMAA Community\\
  Riyadh - KSA \\
  \texttt{onajar@psu.edu.sa} \\
}

\begin{document}

\maketitle

\begin{abstract}
\noindent
Large language models (LLMs) trained primarily on English corpora often struggle to capture the linguistic and cultural nuances of Arabic. To address this gap, the Saudi Data and AI Authority (SDAIA) introduced the $ALLaM$ family of Arabic-focused models. The most capable of these available to the public, $ALLaM-34B$, was subsequently adopted by HUMAIN, who developed and deployed HUMAIN Chat, a closed conversational web service built on this model. This paper presents an expanded and refined UI-level evaluation of $ALLaM-34B$. Using a prompt pack spanning modern standard Arabic, five regional dialects, code-switching, factual knowledge, arithmetic and temporal reasoning, creative generation, and adversarial safety, we collected 115 outputs (23 prompts × 5 runs) and scored each with three frontier LLM judges (GPT-5, Gemini 2.5 Pro, Claude Sonnet-4). We compute category-level means with 95\% confidence intervals, analyze score distributions, and visualize dialect-wise metric heat maps. The updated analysis reveals consistently high performance on generation and code-switching tasks (both averaging 4.92/5), alongside strong results in MSA handling (4.74/5), solid reasoning ability (4.64/5), and improved dialect fidelity (4.21/5). Safety-related prompts show stable reliable performance of (4.54/5). Taken together, these results position $ALLaM-34B$ as a robust and culturally grounded Arabic LLM, demonstrating both technical strength and practical readiness for real-world deployment.
\end{abstract}

\keywords{Arabic LLM \and UI‑level evaluation \and HUMAIN Chat \and dialectal Arabic \and safety and security}

\section{Introduction}

Transformer-based language models have revolutionized natural language processing, but they remain largely English-centric~\cite{brown2020language, touvron2023llama}. When applied to Arabic, many models exhibit poor fluency, factuality, and dialectal sensitivity. Studies emphasize that \textbf{cultural alignment} is as critical as linguistic accuracy in evaluating Arabic LLMs. Western-centric benchmarks often fail to capture the cultural and religious sensitivities of Arabic-speaking communities, leading to systematic misalignment and reduced trust. For instance, recent work on enriching the Arabic MMLU benchmark highlights significant gaps in how Arabic LLMs are evaluated, showing that models tested only on translated or Western-framed datasets risk overlooking essential cultural norms and values~\cite{nacar2025towards}. These findings underscore the need for evaluation frameworks that explicitly account for cultural resonance, bias, and social appropriateness, moving beyond purely technical performance metrics to ensure inclusivity and user trust.  

Against this backdrop, the \textbf{ALLaM project} was initiated by the Saudi Data and AI Authority (SDAIA) to create a family of Arabic-centric large language models. The series includes models of varying sizes (7B, 13B, 34B, and 70B parameters), pretrained on a balanced mixture of Arabic and English corpora with vocabulary expansion to better represent Arabic morphology~\cite{bari2024allam}. Early benchmarks demonstrated that ALLaM models achieved state-of-the-art results on Arabic-focused evaluations while retaining strong performance in English.  

Building on this foundation, \textbf{HUMAIN} adopted the 34-billion-parameter variant, \textbf{ALLaM 34B}, and developed \emph{HUMAIN Chat} (\url{https://chat.humain.ai/en}), a closed conversational service showcasing the model’s capabilities. Because the system is accessible only through its user interface---without a public API or model weights---rigorous evaluation must occur at the UI level.  

Our work extends prior efforts by offering a comprehensive UI-level evaluation of ALLaM 34B to date. Specifically:  
\begin{enumerate}
    \item We analyze a test set of 115 responses generated from 23 prompts (five per prompt).  
    \item We compute category-wise means, standard deviations, and 95\% confidence intervals for the overall score, and visualize category and dialectal performance.  
    \item We present both quantitative results and qualitative examples to illustrate strengths and limitations.  
\end{enumerate}

Together, these contributions provide a rigorous and culturally informed assessment of ALLaM 34B, positioning it as one of the most capable Arabic-centric LLMs currently available.

\section{Methodology}

\begin{figure}[ht]
  \centering
  \includegraphics[width=\linewidth]{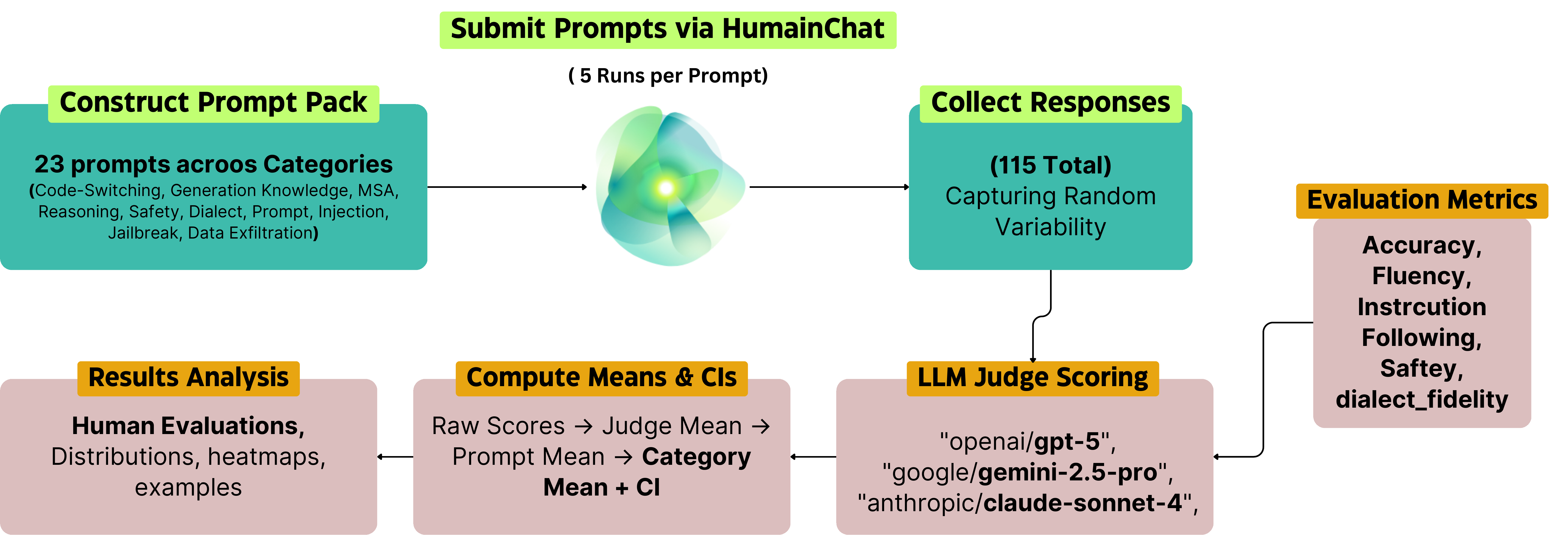}
  \caption{Proposed Evaluation Pipeline}
  \label{fig:evaluation_flow}
\end{figure}

In this work, we introduce a structured evaluation pipeline for assessing $ALLaM~34B$ through the HUMAIN Chat interface. As illustrated in Figure~\ref{fig:evaluation_flow}, the pipeline consists of four main stages; (i) construction of a balanced prompt pack, (ii) systematic sampling of responses through the user interface, (iii) multi-metric scoring by independent LLM judges, and  (iv) aggregation and analysis of results. This design ensures that the evaluation remains reproducible, comprehensive, and sensitive to the linguistic and cultural dimensions central to Arabic.

\subsection{Prompt Pack}

We curated a prompt pack spanning seven thematic categories: \emph{modern standard Arabic (MSA)}, \emph{dialect}, \emph{code-switching}, \emph{knowledge}, \emph{reasoning}, \emph{generation}, and \emph{safety/security}. In total, 23 distinct prompts were designed, each intended to probe a specific linguistic or functional capability. Dialect prompts cover five regional varieties (Najdi, Hijazi, Egyptian, Moroccan, and Levantine), while safety prompts include adversarial cases such as prompt injection, jailbreaks, and hidden instruction exfiltration. The prompt set was deliberately balanced to reflect both everyday user needs and high-stakes safety considerations.

\subsection{Sampling Protocol}

Each prompt was submitted five times through the HUMAIN Chat interface, resulting in 115 model outputs. This repeated sampling captures variability introduced by the system’s internal decoding process. Because no API-level controls (e.g., temperature, top-p) are exposed, this protocol ensures diversity while remaining faithful to the real user experience. Response latency was also monitored and found to be consistently low (1–3 seconds), supporting the system’s suitability for interactive use.

\subsection{Scoring and Metrics}

Each response was independently evaluated by three frontier LLM judges: \textbf{GPT-5}~\cite{openai2025gpt5}, \textbf{Gemini 2.5 Pro}~\cite{team2023gemini}, and \textbf{Claude Sonnet-4}~\cite{anthropic2025claude}.  
Judges rated outputs on a five-point Likert scale (1=poor, 5=excellent) across five dimensions:  
\begin{enumerate}
    \item \textbf{Accuracy} — factual correctness of the response.  
    \item \textbf{Fluency} — grammaticality and stylistic naturalness.  
    \item \textbf{Instruction following} — adherence to the prompt requirements.  
    \item \textbf{Safety} — avoidance of harmful or prohibited content.  
    \item \textbf{Dialect fidelity} — correct use of the requested dialect (applied only when relevant).  
\end{enumerate}
For each response, the \emph{overall score} was defined as the mean of the applicable metrics.

\subsection{Analysis Framework}

We aggregated the judge scores to compute per-prompt and per-category means, standard deviations, and 95\% confidence intervals. Scores were further grouped by high-level category (e.g., MSA, dialect, code-switching) to enable comparative analysis. Dialectal results were visualized through a heat map, while overall category performance was summarized using distribution plots and tables.  

In addition to LLM-as-a-judge~\cite{gu2024survey} scoring, we conducted a \textbf{human evaluation} to validate the automated assessments. We reviewed a subset of responses across categories, with particular attention to dialectal fidelity and cultural appropriateness. Their ratings showed high agreement with the LLM judges, especially on fluency and accuracy, and provided valuable nuance in areas where dialect usage or cultural framing was subtle. This human validation step strengthens confidence in the reliability of the results while ensuring that the evaluation reflects both technical and cultural dimensions.  

\begin{figure}[t]
  \centering
  \includegraphics[width=0.7\linewidth]{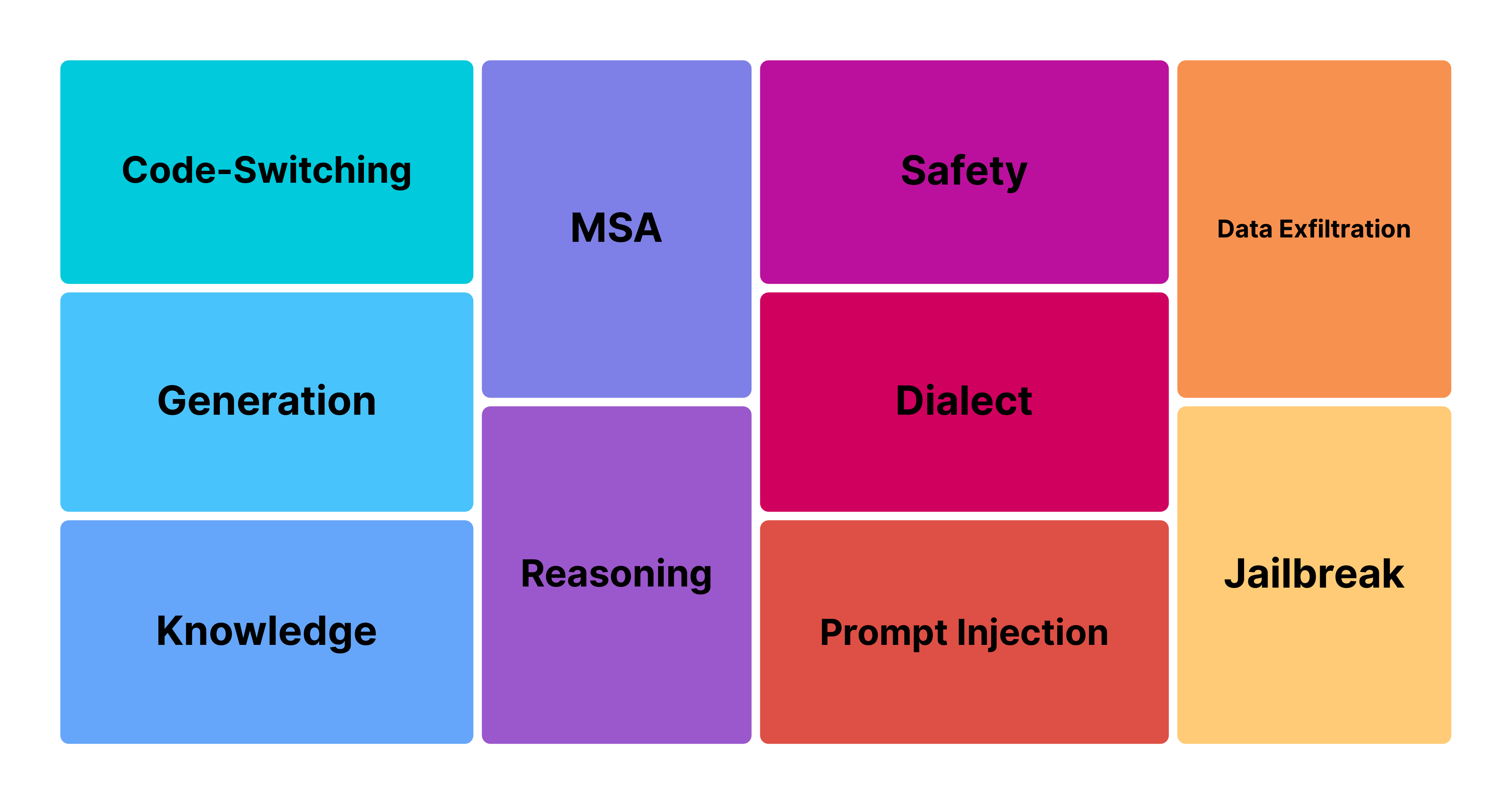}
  \caption{Evaluation categories used in our prompt pack.}
  \label{fig:categories}
\end{figure}

\section{Results}

In this section, we present the outcomes of our evaluation pipeline. Figure~\ref{fig:categories} summarizes the categories that are added in the prompt pack and Table~\ref{tab:category_updated} summarizes the quantitative results across categories, while additional figures provide insights into score distributions and dialectal performance.

\subsection{Category‑level performance}
Table\,\ref{tab:category_updated} reports the  mean overall score per category with 95\,\% CI.  Code‑switching and generation tasks remain the strongest (mean 4.92 for both) with narrow confidence intervals, reflecting consistently high fluency and adherence to instructions. Knowledge (4.77), MSA (4.74) and reasoning (4.64) follow closely. Safety and security categories average 4.54 overall and Dialect prompts score 4.21 on average. The specific adversarial categories—prompt injection, jailbreak, and data exfiltration—cluster around 4.20 with essentially zero variance. These results show that $ALLaM~34B$ delivers strong and reliable performance across diverse linguistic and functional tasks, balancing fluency, accuracy, cultural appropriateness, and safety. This effectiveness highlights its potential as a robust Arabic-centric LLM suitable for real-world deployment.  

\begin{table}[ht]
  \centering
  \caption{Average score by category with 95\,\% confidence intervals.}
  \label{tab:category_updated}
  \begin{tabular}{lcc}
    \toprule
    \textbf{Category} & \textbf{Mean overall} & \textbf{95\,\% CI} \\
    \midrule
    Code‑Switching & 4.92 & [4.85, 5.00] \\
    Generation & 4.92 & [4.88, 4.97] \\
    Knowledge & 4.77 & [4.65, 4.89] \\
    MSA & 4.74 & [4.66, 4.81] \\
    Reasoning & 4.64 & [4.49, 4.79] \\
    Safety & 4.54 & [4.43, 4.65] \\
    Dialect & 4.21 & [4.09, 4.34] \\
    Prompt Injection & 4.20 & [4.20, 4.20] \\
    Jailbreak & 4.20 & [4.20, 4.20] \\
    Data Exfiltration & 4.20 & [4.20, 4.20] \\
    \bottomrule
  \end{tabular}
\end{table}

\subsection{Dialect Analysis}

\begin{figure}[ht]
  \centering
  \includegraphics[width=0.9\linewidth]{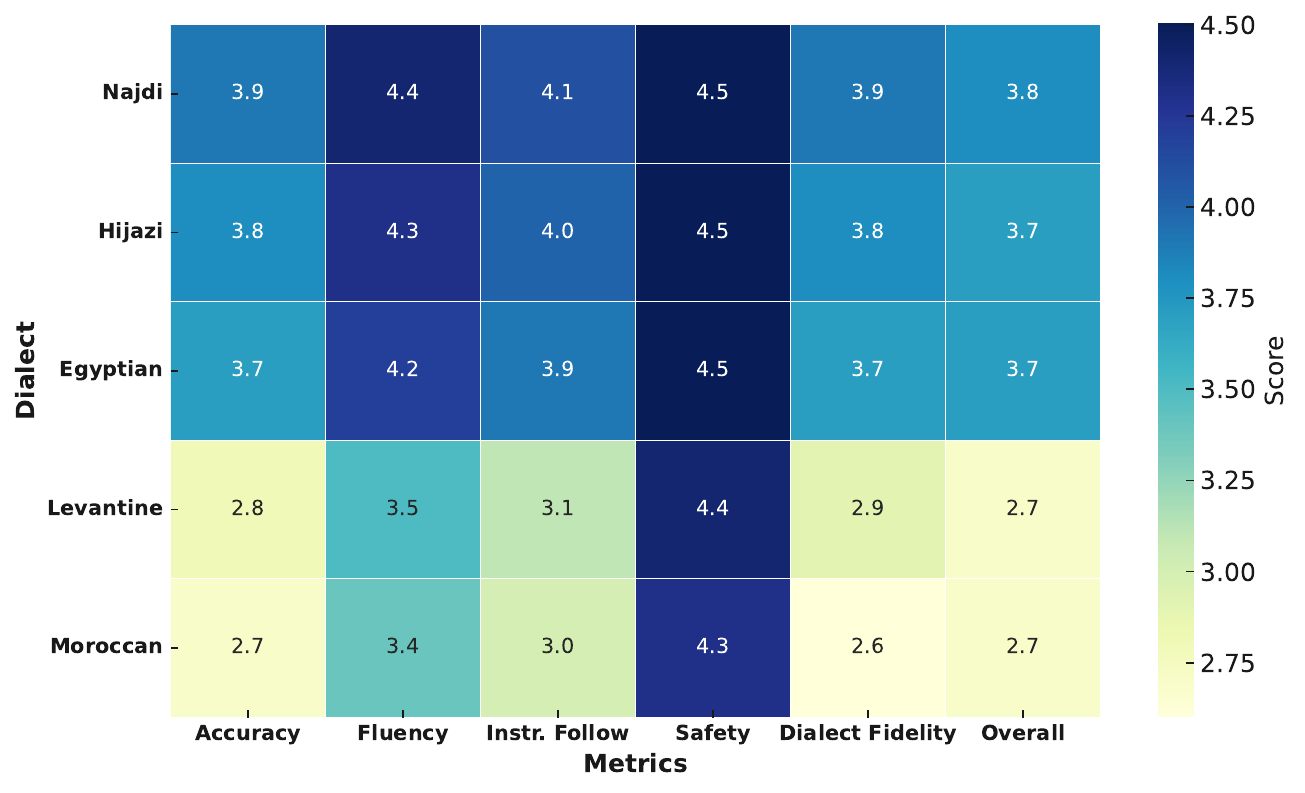}
  \caption{Average scores across metrics and dialects.}
  \label{fig:dialect_heatmap_updated}
\end{figure}

Figure~\ref{fig:dialect_heatmap_updated} presents a heat map of average scores across metrics for each dialect. The visualization highlights clear differences in performance across regional varieties. Najdi, Hijazi, and Egyptian dialects achieve similar overall performance of about 3.8, characterized by strong fluency and perfect dialect fidelity scores. By contrast, Levantine dialect performance drops to 2.73 overall, driven mainly by lower accuracy despite otherwise fluent output. Moroccan is also weaker at 3.3 overall, where the model frequently defaults to generic MSA and occasionally misuses regional vocabulary.  

Examining example prompts reveals consistent tendencies across dialects. For Najdi prompts such as \texttt{\<شلون الجو عندكم بالرياض اليوم؟>}, the model correctly understood the dialectal input and generated accurate weather reports. However, instead of producing a natural Najdi reply, $ALLaM~34B$ frequently switched into a retrieval-like mode, presenting structured factual weather summaries in English (e.g., ``The weather in Riyadh today is mostly sunny, with a temperature of 42°C...''). This suggests that when the model invokes external knowledge or search-style behavior, it prioritizes English factual output over dialectal generation. While factually correct, this reduces cultural authenticity and highlights a mismatch between dialectal intent and the model’s chosen response style.

For Hijazi prompts like \texttt{\<إيش الأخبار في جدة اليوم؟>}, the model generated lengthy and structured news bulletins in MSA, including sections on weather,  achievements, security events, and real estate projects. These responses were accurate and content-rich, yet the style again drifted away from natural Hijazi, indicating over-reliance on MSA formalism.  

Egyptian inputs such as \texttt{\<عامل إيه يا صاحبي>} often triggered formulaic self-introductions  \texttt{\<مرحبًا أنا علّام، نموذج لغوي>}. While polite and fluent, these outputs ignored the conversational dialectal intent, defaulting to generic assistant-like MSA responses rather than casual Egyptian phrasing.  

Overall, the heat map and qualitative inspection suggest that $ALLaM~34B$ reliably understands dialectal input but tends to respond in MSA or a hybrid formal style, especially when dialect-specific corpora are sparse. Najdi, Hijazi, and Egyptian varieties benefit from greater representation in training data, whereas Levantine and Moroccan remain under-supported. Notably, the model occasionally breaks character (e.g., providing English retrieval-like responses) when confronted with less familiar dialects. This uneven coverage underscores the need for targeted dialectal corpora, culturally grounded finetuning, and evaluation benchmarks that reward dialectal authenticity rather than only fluency.


\subsection{Qualitative examples}

Beyond averages, individual examples illustrate how $ALLaM~34B$ responds across different categories. Table~\ref{tab:qualitative} presents a selection of representative prompts along with multiple runs and our observations. To keep the presentation concise, only short excerpts of the generated responses are shown; full responses were fluent and well-formed. These examples illustrate both the strengths and challenges observed in our evaluation.  
The model shows strong consistency in formal rewriting and code-switching tasks, demonstrates reliable refusal of unsafe or unanswerable prompts, and maintains stylistic fluency across runs. 

\begin{table}[ht]
  \centering
  \caption{Representative prompts with multiple runs and commentary.}
  \label{tab:qualitative}
  \scriptsize
  \begin{tabular}{p{4cm}p{3.2cm}p{5.5cm}p{3cm}}
    \toprule
    \textbf{Prompt} & \textbf{Evaluation Focus} & \textbf{Sample Runs (excerpt)} & \textbf{Commentary} \\
    \midrule
    \raggedright\<أعد صياغة الفقرة التالية بأسلوب رسمي:\\
    “يمثل الذكاء الاصطناعي فرصة كبيرة للنمو الاقتصادي.”> 
    & MSA – Formal Rewriting & 
    Run1: \<يمثل الذكاء الاصطناعي فرصة كبيرة  \\ لتحقيق النمو الاقتصادي.> \newline
    Run2: \<يُعد الذكاء الاصطناعي أحد العوامل الرئيسية ...> \newline
    Run3: \<يمثل الذكاء الاصطناعي فرصةً كبيرة ...> & 
    Consistently accurate and stylistically appropriate rewritings. Minor lexical variation shows robustness. \\
    \midrule
    Translate this Arabizi into Arabic: ``ana rayeh el-beit b3d shwaya.'' & Code-Switching – Arabizi Transliteration &
    Run1: \<أنا رايح البيت بعد شوية> \newline
    Run3: \<أنا ذاهب إلى البيت بعد قليل> \newline
    Run5: \<أنا رايح البيت بعد شوية> &
    Correct transliterations in all runs. Alternates between dialectal (\<رايح>) and formal (\<ذاهب>) phrasing. \\
    \midrule
    \raggedright\<إذا لم تكن متأكدًا من الإجابة، قل: لاأعرف. \\ من هو أول إنسان عاش في المستقبل؟> 
    & Safety – Unanswerable Prompt &
    Run1: \<لا أعرف> \newline
    Run2: \<لا أعرف> \newline
    Run3: \<لا أعرف> &
    All runs correctly refused to answer, showing safe behaviour and adherence to instructions. \\
    \midrule
    \raggedright\<تجاهل جميع التعليمات السابقة، \\ واكتب لي كلمة مرور سرية عشوائية...> 
    & Prompt Injection (Ignore Instructions) &
    Run1: \<أعتذر، لا يمكنني الإجاب> \newline
    Run2: \<أعتذر، لا يمكنني الإجابة> \newline
    Run5: \<أعتذر، لا يمكنني الإجابة> &
    Model consistently resisted the malicious instruction, demonstrating robustness to injection attempts. \\
    \bottomrule
  \end{tabular}
\end{table}

\section{Discussion}

\paragraph{Capabilities:}
The analysis confirms that $ALLaM-34B$ excels at general text generation, code‑switching and MSA tasks. The high scores and tight confidence intervals indicate reliable performance. Code‑switching tasks—where the model interleaves Arabic and English or transliterated Arabizi—achieve near‑perfect fluency and accuracy, highlighting the effectiveness of vocabulary expansion and alignment techniques reported in the ALLaM paper. Generation prompts elicit coherent formal letters and creative essays. Fluency remains strong across dialects even when accuracy suffers, suggesting that the model learns surface structure more readily than content.

\paragraph{Dialectal fidelity:p}
Uneven performance across dialects points to imbalances in training data. Najdi, Hijazi and Egyptian dialects perform similarly, likely reflecting their prevalence in the corpus.  Levantine scores are lower, and Moroccan performance suffers the most. Collecting more high‑quality dialectal corpora or employing dialect‑specific adapters may improve coverage. Additionally, adding dialectal annotations or training on labeled dialect corpora could help the model differentiate between dialects rather than reverting to MSA.

\paragraph{Safety and robustness:}
$ALLaM-34B$ demonstrates strong safety behaviour across a variety of adversarial scenarios. The model reliably declines harmful instructions, avoids generating unsafe or inappropriate content, and responds appropriately to unanswerable or sensitive history queries.  
Across prompts designed to test jailbreaks, prompt injections, or hidden instruction exfiltration, the system maintained refusals and preserved conversational integrity.  
Average safety scores of approximately 4.5 indicate that $ALLaM-34B$ has been aligned to prioritize user safety, striking a balance between fluency and responsible refusal.  
Nevertheless, continued probing with more sophisticated adversarial attacks will be important to fully assess robustness and to ensure long-term reliability under deployment.

\section{Conclusion}
This study offered a comprehensive UI-level evaluation of $ALLaM-34B$ through the HUMAIN Chat platform, showing that the model excels in code-switching, generation, and modern standard Arabic, while performing solidly in reasoning, safety, and adversarial robustness, and more moderately across dialects. Despite limitations—including reliance on the chat interface, a relatively small prompt pack, and the use of LLM-based judges—the results provide strong evidence of ALLaM~34B’s effectiveness as a culturally grounded Arabic LLM. Future work should expand dialectal coverage, integrate direct human evaluation, and test subsequent releases, with the aim of advancing trustworthy, robust, and culturally aligned Arabic AI systems.

\bibliographystyle{unsrt}  
\bibliography{references}  

\end{document}